\documentclass{article}
\usepackage{ijcai18}

\usepackage{times}
\usepackage{xcolor}
\usepackage{soul}
\usepackage[utf8]{inputenc}
\usepackage[small]{caption}
\usepackage{bm}

\usepackage{epsfig}
\usepackage{graphicx}
\usepackage{amsmath}
\usepackage{amssymb}
\usepackage[utf8]{inputenc} 
\usepackage[T1]{fontenc}    
\usepackage{booktabs}       
\usepackage{amsfonts}       
\usepackage{nicefrac}       
\usepackage{microtype}      
\usepackage{multirow}

\usepackage{fancyhdr}

\usepackage{tabularx}
\usepackage{color}
\usepackage{bm}
\usepackage{algorithm}
\usepackage{algorithmic}
\usepackage{multirow}
\usepackage{subfigure}
\usepackage{amsthm}
\newtheorem{thm}{Theorem}
\newtheorem{cor}{Corollary}
\newtheorem{prop}{Proposition}
\newtheorem{assumption}{Assumption}

\newtheorem{remark}{Remark}
\newtheorem{condition}{Condition}

\title{Toward Designing Convergent Deep Operator Splitting Methods \\ for Task-specific Nonconvex Optimization}

\author{
	Risheng Liu$^{1,2}$\thanks{Corresponding Author.},
	Shichao Cheng$^{2,3}$, 
	Yi He$^{1,2}$, 
	Xin Fan$^{1,2}$,
	Zhongxuan Luo$^{1,2,3}$
	\\
	$^1$International School of Information Science \& Engineering, Dalian University of Technology\\
	$^2$Key Laboratory for Ubiquitous Network and Service Software of Liaoning Province\\
	$^3$School of Mathematical Science, Dalian University of Technology\\ 
	\{rsliu, xin.fan, zxluo\}@dlut.edu.cn,
	\{shichao.cheng, heyiking\}@outlook.com
}

\begin{document}

\maketitle
\thispagestyle{fancy}
\fancyhead{}
\chead{}
\rhead{}
\lfoot{}
\cfoot{\thepage}   
\rfoot{}
\fancyhead[C]{2018 International Joint Conferences on Artificial Intelligence}
\renewcommand{\headrulewidth}{0pt}
\renewcommand{\footrulewidth}{0.7pt}
\begin{abstract}
	Operator splitting methods have been successfully used in computational sciences, statistics, learning and vision areas to reduce complex problems into a series of simpler subproblems. However, prevalent splitting schemes are mostly established only based on the mathematical properties of some general optimization models. So it is a laborious process and often requires many iterations of ideation and validation to obtain practical and task-specific optimal solutions, especially for nonconvex problems in real-world scenarios. To break through the above limits, we introduce a new algorithmic framework, called Learnable Bregman Splitting (LBS), to perform deep-architecture-based operator splitting for nonconvex optimization based on specific task model. Thanks to the data-dependent (i.e., learnable) nature, our LBS can not only speed up the convergence, but also avoid unwanted trivial solutions for real-world tasks. Though with inexact deep iterations, we can still establish the global convergence and estimate the asymptotic convergence rate of LBS only by enforcing some fairly loose assumptions. Extensive experiments on different applications (e.g., image completion and deblurring) verify our theoretical results and show the superiority of LBS against existing methods.		
\end{abstract}

\section{Introduction}

In this work, we consider the optimization problem
\begin{equation}
\min\limits_{x_n\in\mathcal{X}_n}\Psi(x):=f(x) + \sum_{n=1}^N g_n(x_n),\label{eq:model}
\end{equation}
where $x=\{x_1,\cdots,x_N\}\in\mathbb{R}^D$ has $N$ blocks ($N\geq 1$), $f(x)$ is continuously differentiable, but a series of $g_n(x)$ is not necessarily differentiable. Notice that convexity is not assumed for $f$, $g_n$ or $\mathcal{X}_n$. By considering $g_n$ as an extended value function (i.e., take $+\infty$ value), we can incorporate the set constraint $x_n\in\mathcal{X}_n$ into $g_n(x_n)$ since this is equivalent to minimize the indicator function of $\mathcal{X}_n$. Therefore, we will not include this set constraint in our following analysis. Typically, $f$ captures the loss of data fitting from the specific task modeling and $g=\sum_{n=1}^Ng_n$ is the regularization that promotes desired structures on the variable $x$.
Problems appearing in many learning and vision applications, such as sparse coding~\cite{lin2011linearized}, tensor factorization~\cite{xu2017globally}, image restoration~\cite{liu2018proximal}, and data clustering~\cite{yang2016ell}, can all be (re)formulated in the form of Eq.~\eqref{eq:model}.

\subsection{Related Works}

One of the most prevalent algorithms to solve Eq.~\eqref{eq:model} is the operator splitting  approach. The main idea behind such kind of schemes is to reduce complex problems built from simple pieces into a series smaller subproblems which can be solved sequentially or in parallel. 
In the past several decades, a variety of splitting methods have been designed and analyzed. For example, \cite{passty1979ergodic} provided a prototype of the Forward-Backward Splitting (FBS) and proved its ergodic convergence. \cite{beck2009a} presented the convergence rate of proximal gradient (PG) and accelerated proximal gradient (APG, also known as FISTA). Recently, \cite{davis2016convergence} provided a unified way to analyze the convergence rate of Peaceman-Rachford Splitting (PRS) and Douglas-Rachford Splitting (DRS). It is also known that the widely used Alternating Direction Method of Multipliers (ADMM) can be reformulated within the operator splitting (e.g., DRS) framework in the dual space~\cite{lin2015linearized}. Though with mathematically proved convergence properties, the generally designed algorithms may still fail on some particular nonconvex optimization models in real scenarios. This is mainly because that due to their fixed updating schemes, it is hard to escape the unwanted saddle points during iterations.

To improve the performance in practical real-world applications, some researches tried to parameterize exiting iteration schemes and learned their parameters in the resulted propagation models. For example, \cite{Liu2016Learning} learned the parameters of a parameterized partial differential equation for various image and video processing tasks. Similarly, \cite{chen2015learning} introduced a higher-order diffusion system to perform data-dependent gradient descent for image denoising and super-resolution. The studies in \cite{schmidt2014shrinkage} and \cite{Yang2017ADMM} respectively parametrized the half-quadratic splitting and ADMM for practical applications, such as non-blind deconvolution and MRI imaging. Very recently, inspired by the success of deep networks in different application fields, some works also tried to replace the standard iterations by existing network architectures. By considering convolutional neural networks (CNNs) as special image priors, \cite{zhang2017learning} proposed an iterative CNN scheme to address image restoration problems. 

However, we have to point out that although with relatively better performance in some specific tasks, the nice convergence properties proved from theoretical side are completely missing in these methods. That is, neither the adaptive parameterization nor the replaced CNNs mentioned above can preserve the convergence results proved for the original iteration schemes. 
Moreover, it is even impossible to investigate and control the iterative behaviors (e.g., descent) of these methods, since their learned iterations actually no longer solve the original optimization model.

\subsection{Contributions}

In this work, we propose Learnable Bregman Splitting (LBS), a novel deep operator splitting algorithm for nonconvex optimization in real-world scenarios. Specifically, we first introduce a Bregman distance function to penalize the variables at each iteration. Then the basic LBS updating scheme is established based on a relaxed Krasonselskii-Mann iteration~\cite{davis2016convergence}. By introducing a novel triple operator splitting strategy, we can successfully combine the task-model-inspired and data-learning-driven operators within the LBS algorithmic framework. In summary, our contributions mainly include:
\begin{itemize}
	\item LBS provides a novel learning strategy to extend prevalent mathematically designed operator splitting schemes for task-specific nonconvex optimization. Thanks to the learnable deep architectures, we can learn our iterations on collected training data to avoid unwanted solutions in particular applications.  
	\item Different from most existing learning-based optimization algorithms (e.g., iteration parameterization and CNN incorporation methods mentioned above), in which there is no theoretical guarantee, we provide rich investigations on the iterative behaviors, prove the global convergence and estimate the convergence rate of our LBS.
	\item We also demonstrate how to apply our algorithm for different computer vision applications and extensive results verify that LBS outperforms state-of-the-art methods on all the compared problems.
\end{itemize}

\section{Learnable Bregman Splitting Method}

In this section, a learning-based operator splitting method, named Learnable Bregman Splitting (LBS), is developed for the nonconvex optimization model in Eq.~\eqref{eq:model}.

\subsection{Bregman Distance Penalization}
As a fundamental proximity measure, the Bregman distance\footnote{The use of Bregman distance in optimization within various contexts is well spread. Many interesting properties of this function can be found in the comprehensive work \cite{bauschke1997legendre}.} plays important roles in various iteration algorithms. However, since it does not satisfy the triangle inequality nor symmetry, this function is not a real metric.  
Given a convex differential function $h$, the associated Bregman distance can be written as
\begin{equation}
\Delta_h(x,y)=h(x)-h(y)-\langle\nabla h(y), x-y\rangle.
\end{equation}
Clearly, $\Delta_h$ is strictly convex with respect to the first argument. Moreover, $\Delta_h\geq 0$ for all $(x,y)$ and is equal to zero if and only if $x=y$. So $\Delta_h$ actually provides a natural (asymmetric) proximity measure between points in the domain of $h$. 

In this work, we introduce $\Delta_{h_n}$ as a penalty term for each $x_n$ at $t$-th iteration. That is, we actually minimize the following energy to update $x^{t+1}$:
\begin{equation}
\Psi_{h}^{t+1}(x)=f(x) + \sum\limits_{n=1}^N g_n(x_n) + \frac{1}{\lambda}\Delta_{h}(x,x^t),\label{eq:psi_d}
\end{equation}
where we denote $\Delta_{h}(x,x^t)=\sum_{n=1}^N\Delta_{h_n}(x_n,x_n^t)$ and $\lambda>0$ is the penalty parameter. It will be demonstrated that
$\{\Delta_{h_n}(x_n,x_n^t)\}$ brings nice convergence properties for the proposed optimization model when it is $\mu$-strong convex~\cite{bauschke1997legendre}.

\subsection{Uniform Coordinate Updating Scheme}	
In this work, we consider the following general coordinate update scheme to minimize the energy function in Eq.~\eqref{eq:psi_d}:
\begin{equation}
x_n^{t+1}=x_n^t - \gamma^t[\mathcal{D}(x^t)]_n,\label{eq:cd} 
\end{equation}
where $\mathcal{D}(x^t)$ denotes the update direction (regarding to the problem) on $x^t$, $\gamma^t>0$ is a step size and $[\cdot]_n$ denotes the $n$-th block of the given variable. 
It should be pointed out that by formulating  $\mathcal{D}:=\mathcal{I}-\mathcal{T}$ (here $\mathcal{I}$ denotes the identity mapping), Eq.~\eqref{eq:cd} can be further recognized as a relaxed Krasonselskii-Mann iteration \cite{shi2016primer} with the operator $\mathcal{T}$ (i.e.,
$x_n^{t+1}=(1-\gamma^t)x_n^t + \gamma^t[\mathcal{T}(x^t)]_n$) and then various existing first-order schemes can be reformulated in the form of Eq.~\eqref{eq:cd}. 

Specifically, by defining $\mathcal{J}_{\mathcal{F}}=(\mathcal{I}-\mathcal{F})^{-1}$ (resolvent),  $\mathcal{R}_{\mathcal{F}}=2\mathcal{J}_{\mathcal{F}}-\mathcal{I}$ (reflection) for $\mathcal{F}$ (operator about $f$), we can obtain a variety of prevalent splitting schemes, such as FBS, PRS, and DRS. As for the operator $\mathcal{T}$ in our work, if setting $\mathcal{T}=\mathcal{I}-\mathcal{J}_{\mathcal{G}}\circ(\mathcal{I}-\mathcal{F})$ and $\gamma^t=1$, we obtain FBS from Eq.~\eqref{eq:cd}, i.e.,
$x^{t+1} = \mathcal{J}_{\mathcal{G}}\circ(\mathcal{I}-\mathcal{F})(x^{t}),$ 
where $\circ$ denotes the operator composition. 
By considering $\mathcal{J}_{\mathcal{G}}=\mathtt{prox}_{g}$\footnote{The proximal operation with respect to $g$ (denoted as $\mathtt{prox}_{g}$) is defined as $\mathtt{prox}_{g}(x): =\arg\min_{y} g(y)+ \frac{1}{2}\|y-x\|^{2}$.} and   $\mathcal{F}=\nabla f$, we further have the well-known  proximal (or projected) gradient scheme from FBS. 
Setting $\mathcal{T} = \mathcal{R}_{\mathcal{G}}\circ\mathcal{R}_{\mathcal{F}}$ and $\gamma^{t}=1$, Eq.~\eqref{eq:cd} reduces to
$x^{t+1} = \left(\mathcal{R}_{\mathcal{G}}\circ\mathcal{R}_{\mathcal{F}}\right)(x^{t}),$
which is just the standard PRS iteration. 
Similarly, with the same $\mathcal{T}$ in PRS and $\gamma = 1/2$, we can also deduce DRS.

Additionally, it should be pointed out that the well-known ADMM \cite{lin2011linearized} can also be deduced by applying DRS on its Lagrange dual space \cite{davis2016convergence}. Therefore, although the original ADMM is designed for linearly constraint models, we can still reformulate it as a special case of Eq.~\eqref{eq:cd} in the dual variable space. Thus Eq.~\eqref{eq:cd} actually can also be utilized to address the  constrained problems.

In consideration of the nice properties and high flexibility of Eq.~\eqref{eq:cd} mentioned above, we would like to utilize the general updating scheme in Eq.~\eqref{eq:cd} with $\mathcal{D}=\mathcal{I}-\mathcal{T}$, (named Uniform Coordinate Updating Scheme, UCUS for short) as our fundamental block-wise iteration rule. 

\subsection{Splitting with Learnable Architecture}
As discussed above, most existing splitting algorithms (e.g., FBS, PRS and DRS) specify the operator only based on the optimization model. However, due to the nonconvex nature of the model, it is hard for these schemes to escape undesired local minimum. Moreover, the complex data distribution in real applications will also slow down the redesigned iterations.

To partially address these issues, we provide a new splitting strategy, in which a learnable operator $\mathcal{T}_d$ is introduced to extract information from the data. That is, we consider the following triple splitting scheme:
\begin{equation}
[\mathcal{T}(x)]_n:=\mathcal{T}_{g_n}\circ\mathcal{T}_{f_{\lambda}}\circ\mathcal{T}_d(x), \ (n=1,2,\cdots,N),
\end{equation}
where $\mathcal{T}_{g_n}$ and $\mathcal{T}_{f_{\lambda}}$ are operators related to $\Psi_{h}^t$ in Eq.~\eqref{eq:psi_d}. Here we just follow a FBS-like strategy to define
$\mathcal{T}_{f_{\lambda}}=\mathcal{I}-\nabla(f+\frac{1}{\lambda}\Delta_h)$ and $\mathcal{T}_{g_n}=\mathtt{prox}_{\rho g_n}$. As for $\mathcal{T}_d$, we would like to build it as a learnable network architecture and train its parameters from collected training data set\footnote{See Sec.~\ref{sec:exp} for the details of this operator and its training strategy.}. In this way, we can successfully incorporate data information to improve the iterative performance of the proposed algorithm. 

Notice that it is challenging to analyze the convergence issues for the existing network-incorporated iterations (e.g., \cite{zhang2017learning}), since all their schemes are built in heuristic manners. \emph{In contrast, we will demonstrate in the following section that the convergence of our LBS can be strictly proved.}

\subsection{The Complete Algorithm}

It can be seen that the learnable operator $\mathcal{T}_d$ are not deduced from strict optimization rule, there may exists iteration errors when calculating $\mathcal{T}$ at each stage. Thus we introduce a new condition to control the inexactness of our updating scheme at each iteration. Specifically, we define the optimality errors of a given variable $u$ at $t$-th iteration based on the first order subdifferential of $\Psi_h^{t+1}$, i.e., 
$$
\begin{array}{l}
e_{u_n}^{t+1}:=d_{g_n} + \nabla_n f(\{x_{n^-}^{t+1},u_n,x_{n^+}^{t}\}) \\
\qquad + \frac{1}{\lambda^t}(\nabla h_n(u_n)- \nabla h_n(x_n^t)),
\end{array}
$$
where $d_{g_n}\in\partial g_n(u_n)$ (here we denote $\partial g_n$ as the limiting Ferchet subdifferential of $g_n$ \cite{xu2017globally}) and $n=1,2,\cdots,N$. Then we consider the following so-called Relaxed Optimality Condition (ROC) for the given $u\in\mathbb{R}^D$.
\begin{condition}(Relaxed Optimality Condition) Given any $u\in\mathbb{R}^D$, we define the relaxed optimality condition of $\Psi_h^{t+1}$ for $u_n\in u$ ($n=1,2,\cdots,N$) as 
	$\|e_{u_n}^{t+1}\|\leq c\|u_n-x_n^t\|$, where $c$ is a fixed positive constant.
\end{condition}

Based on the above condition, we are ready to propose our LBS algorithm for solving Eq.~\eqref{eq:model} in Alg.~\ref{alg:LBS}.
Notice that the UCUS iteration, denoted as $\mathtt{ucus}(\cdot)$, are independently stated in Alg.~\ref{alg:ucus}. It can be seen that if ROC is satisfied, the LBS iterations are fully based on the learnable network operator. While for some iterations, which do not satisfy ROC, we may still perform the model-based operators $\mathcal{T}_{g_n}\circ\mathcal{T}_{f_{\lambda^t}}$ to guarantee the final convergence. For convenience, hereafter the subvectors $\{x_1, \dots, x_{n-1}\}$ and $\{x_{n+1},\dots,x_N\}$ are denoted as $x_{n^-}$ and $ x_{n^+}$ for short, respectively. We also denote $\psi_{n}^{t+1}(x_n) = f_n^{t+1}(x_n) + g_n(x_n)$, in which $f_n^{t+1}(x_n)=f(\{x_{n-}^{t+1}, x_n, x_{n+}^{t}\})$. 

\begin{algorithm}
	\caption{Learnable Bregman Splitting (LBS)}\label{alg:LBS}
	\begin{algorithmic}[1]
		\REQUIRE $x^0$, $\mathcal{T}_d$, $\Delta_h$, $c$, $\mu$, $\{\gamma^t | 0<\gamma^{t}\leq 1\}$, $\{\lambda^t|\lambda^t>0\}$.
		\WHILE{not converged} 
		\STATE $z^{t+1}=\mathcal{T}_{f_{\lambda^t}}\circ\mathcal{T}_d(x^{t})$.
		\FOR {$n = 1, 2,\dots,N$}
		\STATE $u^{t+1}_n =\mathcal{T}_{g_n}(z_n^{t+1})$.\label{step:c-error}
		\IF {$u_n^{t+1}$ satisfies ROC}
		\STATE $v^{t+1}_n = u^{t+1}_n$.
		\ELSE
		\STATE $v^{t+1}_n = \mathcal{T}_{g_n}\circ\mathcal{T}_{f_{\lambda^t}}(\{x^{t+1}_{n^-}, x_n^{t}, x^{t}_{n^+}\})$.\label{step:proximal}
		\ENDIF
		\STATE $x_n^{t+1}=\mathtt{ucus}(\{x^{t+1}_{n^-}, v_n^{t+1}, x^{t}_{n^+}\}, \gamma^t)$.
		\ENDFOR	
		\ENDWHILE
	\end{algorithmic}
\end{algorithm}

\begin{algorithm}
	\caption{$x_n^{t+1}=\mathtt{ucus}(\{x^{t+1}_{n^-}, v_n^{t+1}, x^{t}_{n^+}\}, \gamma^t)$}\label{alg:ucus}
	\begin{algorithmic}[1]
		\STATE $w^{t+1}_n = x^{t}_n - \gamma^{t}(x^{t}_n - v^{t+1}_n)$.
		\IF {$\psi_n^{t+1}(w^{t+1}_n) \leq \psi_n^{t+1}(v^{t+1}_n)$ } \label{step:function-com}
		\STATE $x^{t+1}_n = w^{t+1}_n$.
		\ELSE
		\STATE $x^{t+1}_n = v^{t+1}_{n}$.
		\ENDIF
	\end{algorithmic}
\end{algorithm}

\section{Convergence Analysis}
In this section, we provide strict analysis on the convergence behaviors of LBS. 
The following assumptions on the functions $f$, $g$, and $\Psi$ are necessary for our analysis. Notice that all these assumptions are fairly loose in optimization area and satisfied in most vision and learning problems.
\begin{assumption}\label{ass:function}
	1) $f$ is Lipschitz smooth and $g_n$ is proximable\footnote{A function $g$ is proximable if it is easy to obtain the minimizer of $g(x) + \frac{1}{2\beta}\|x-y\|$ for any given y and $\beta>0$.}. 
	2) $\Psi$ is coercive. 
\end{assumption}

The roadmap of our analysis is summarized as follows: We first prove that the non-increase of objective, the boundedness of the variables sequence, and the convergence of subsequence in Propositions ~\ref{prop:Non-increasing}, \ref{prop:finit-sum}, and \ref{prop:sub-convergence}, respectively.
Then prove Theorem~\ref{thm:lbs} that LBS can generate Cauchy sequences, which converge to the critical points of the model in Eq.~\eqref{eq:model}. The convergence rate of the sequences is also analyzed in Corollary~\ref{cor:lbs}. The detailed proofs are represented on the arXiv report ().


\begin{prop}\label{prop:Non-increasing}
	(Sufficient descent). 
	If $c < \frac{\mu}{2\lambda}$ and $\rho < \frac{1}{L}$, both the learnable operators $\mathcal{T}_{g_n} \circ \mathcal{T}_{f_{\lambda}} \circ\mathcal{T}_d$ and $\mathcal{T}_{g_n}\circ\mathcal{T}_{f_{\lambda}}$ in Alg.~\ref{alg:LBS} can get the objective inequality: 
	$$\psi_{n}^{t+1}\left(v_n^{t+1}\right)\leq\psi_{n}^{t+1}\left(x_n^t\right)-M \|u^t-x^t\|^2,$$
	where $M = \max\{ \frac{\mu}{2\lambda} - c, \frac{1}{2 \rho} - \frac{L}{2}\}$, $L$ is Lipschitz moduli of $\nabla f$. 
	Together with the direct comparison of function values in Alg.~\ref{alg:ucus}, there exists a non-increasing objective sequence, i.e.,
	$$
	\begin{array}{l}
	\qquad\psi_{n}^{t+1}(x_{n}^{t+1}) \leq \psi_{n}^{t+1}(v_n^{t+1}) \leq \psi_{n}^{t+1}(x_n^{t}).
	\end{array}
	$$
\end{prop}

\begin{remark}
	The inequalities in Proposition~\ref{prop:Non-increasing} builds the relationship of $\psi_{n}^{t+1}(x_n^t)$ and $\psi_{n}^{t+1}(v_n^{t+1})$, thus we can obtain a series of useful inequalities:
	$$
	\begin{array}{l}
	\quad\Psi(x^{t+1}) = \Psi(x_{N+}^{t+1})\\
	\leq \Psi(\{x_{n-}^{t+1},v_n^{t+1},x_{n+}^{t}\}) \leq \Psi(\{x_{n-}^{t+1},x_n^{t+1},x_{n+}^{t}\}) \\
	\leq \Psi(\{v_1^{t+1},x_{1+}^{t}\}) \leq \Psi(\{x_1^{t},x_{1+}^{t}\}) = \Psi(x^{t}),
	\end{array}
	$$
	where $n=N-1, \dots, 1$. It implies the non-increasing property of $\{\Psi(x^t)\}_{t\in\mathbb{N}}$.
\end{remark}

\begin{prop}\label{prop:finit-sum}
	(Square summable).
	If $\gamma \in (0,1]$, $\{x^{t}\}_{t\in\mathbb{N}}$, $\{v_n^{t}\}_{t\in\mathbb{N}}$ are the sequences by Alg.~\ref{alg:LBS}, we have 
	$$
	\sum\limits_{t=1}^{\infty}\|x^{t+1}-x^{t}\|^{2} \leq \sum\limits_{t=1}^{\infty} \sum\limits_{n=1}^{N} \|v_n^{t+1} - x_n^t\|^{2} < \infty.
	$$
\end{prop}

\begin{prop}\label{prop:sub-convergence}
	(Subsequence convergence). Let $\{x^t\}_{t\in\mathbb{N}}$ be the sequence generalized by Alg.~\ref{alg:LBS}. 
	If $x^*$ is any accumulation point of $\{x^t\}_{t\in\mathbb{N}}$. Then we have
	$$
	x_n^{t_j} \to x^*, v_n^{t_j} \to x^{*}, \ 
	\lim_{j \to \infty}\Psi(v^{t_j}) = \Psi(x^*),
	$$
	when $j \to \infty$.
\end{prop}

\begin{remark}
	Indeed, Propositions~\ref{prop:finit-sum} and \ref{prop:sub-convergence} are the key points to prove that the sequence $\{x^t\}_{t\in\mathbb{N}}$ has critical points. Proposition~\ref{prop:sub-convergence} can be derived by combining the Lipschitz smoothness of $f$, lower semi-continuous of $g_n$, $\mu$-strong convexity of $h$, and property of the learnable operators $\mathcal{T}_{g_n} \circ \mathcal{T}_{f_{\lambda}} \circ\mathcal{T}_d$ and $\mathcal{T}_{g_n}\circ\mathcal{T}_{f_{\lambda}}$ in Steps~\ref{step:c-error} and \ref{step:proximal} of Alg.~\ref{alg:LBS}. 
\end{remark}

\begin{thm}\label{thm:lbs}
	(Critical point and Cauchy sequence).
	Let Assumption~\ref{ass:function} hold for Eq.~\eqref{eq:model},
	then the sequences $\{x^t\}_{t\in\mathbb{N}}$ generated by Alg.~\ref{alg:LBS} has critical points $x^*$ of $\Psi$, 
	i.e., if $\Psi^*$ is the limit of sequence $\{\Psi(x^{t})\}_{t\in\mathbb{N}}$, We have 
	$$\Psi(x^*) = \Psi^*, 0 \in \partial \Psi(x^{*}).$$
	If $\Psi$ is a Kurdyka–Łojasiewicz function\footnote{It should be pointed out that many functions arising in learning and vision areas, including $\ell_0$ norm and rational $\ell_p$
		norms (i.e., $p=p_1/p_2$) are all Kurdyka–Łojasiewicz functions~\cite{lin2015linearized}.}, we can further prove that $\{x^t\}_{t\in\mathbb{N}}$ is a Cauchy sequence, thus globally converges to a critical point of $\Psi$.
\end{thm}
Based on the above theorem, we can estimate convergence rate as follows.
\begin{cor}\label{cor:lbs}
	Let $\phi(s)=\frac{t}{\theta}(s)^{\theta}$ be a desingularizing function with a constant $t>0$ and a parameter $\theta\in(0,1]$. Then $\{x^t\}_{t\in\mathbb{N}}$ generated by Alg.~\ref{alg:LBS} converges after finite iterations if $\theta=1$. The linear and sub-linear rates can be obtained if choosing $\theta\in[1/2,1)$ and $\theta\in(0,1/2)$, respectively.
\end{cor}

\begin{figure}[h]
	\centering
	\begin{tabular}{c@{\extracolsep{0.1em}}c@{\extracolsep{0.1em}}c@{\extracolsep{0.1em}}c@{\extracolsep{0.1em}}c}
		\multicolumn{5}{c}{\includegraphics[width=0.45\textwidth]{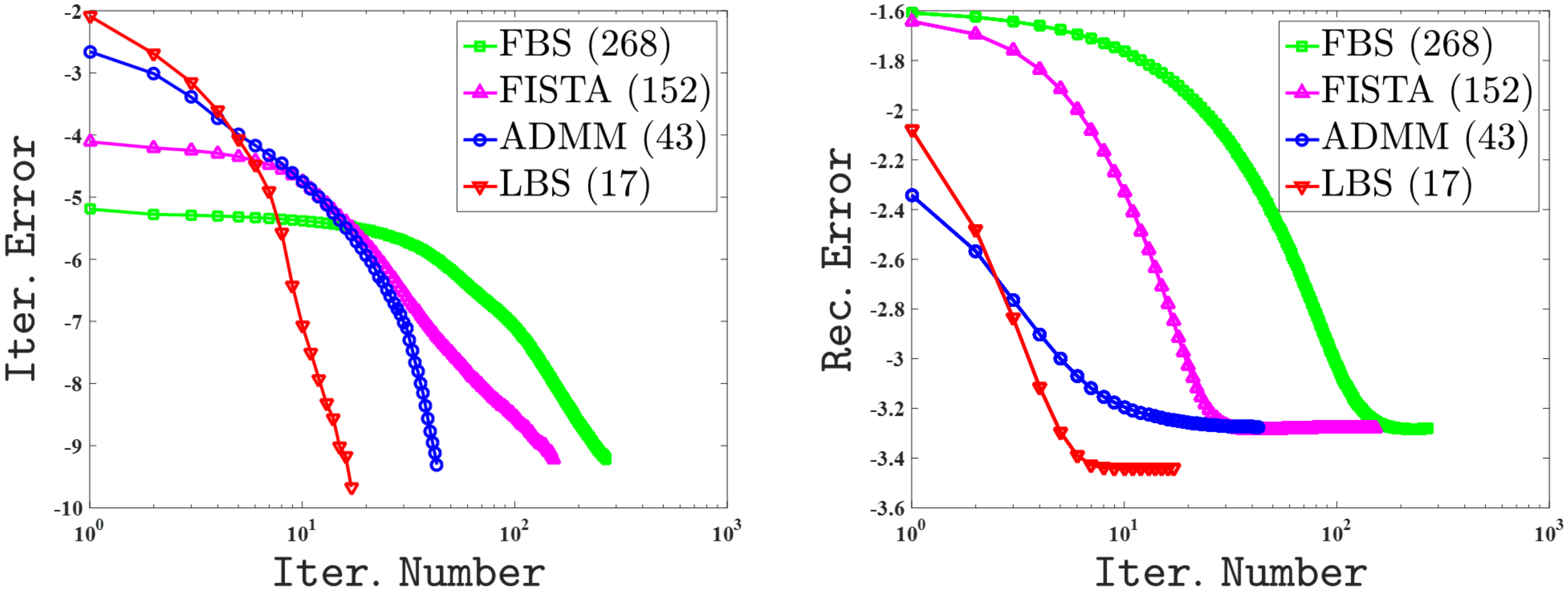}}\\
		\includegraphics[width=0.09\textwidth]{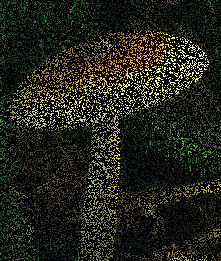}
		&\includegraphics[width=0.09\textwidth]{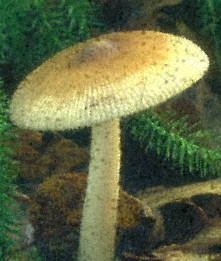}
		&\includegraphics[width=0.09\textwidth]{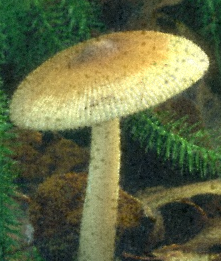}
		&\includegraphics[width=0.09\textwidth]{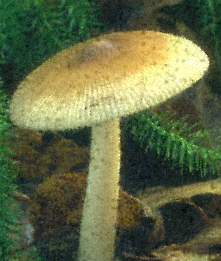}
		&\includegraphics[width=0.09\textwidth]{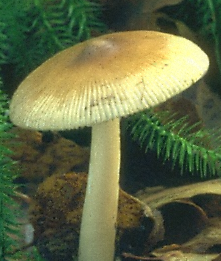}\\
		Input  & FBS &FISTA  & ADMM & LBS \\	
	\end{tabular}
	\caption{The iteration behaviors and visual results of LBS with comparisons to various splitting approaches, including FBS, FISTA, and ADMM. We compare the ``$\mathtt{Iter.~ Error}$'' and ``$\mathtt{Rec. ~Error}$'' of them on the top row and indicate the complete number of iterations in brackets after each method in the legend. The second row is the final restored results of all compared methods. The quantitative scores (PSNR / SSIM) of these methods are 25.27 / 0.65 (FBS), 25.36 / 0.65 (FISTA), 25.25 / 0.64 (ADMM), 28.45 / 0.83 (LBS), respectively.}
	\label{fig:convergence}
\end{figure}
\section{Numerical Results}\label{sec:exp}
To verify the convergence and performance of LBS for nonconvex optimization, we apply it on two widely researched vision problems, i.e., image completion and deblurring. In our algorithm, we adopt residual network as the learnable network architecture for $\mathcal{T}_d$, which can well describes the sparse priors. Specially, there are 19 layers in our network which includes 7 convolution layers, 6 ReLU layers, 5 batch normalization layers and one loss layer.
Every convolution layer has 64 kernels of size $3\times3$, and possesses the dilation attribute.  
In training stage, we randomly select 800 natural images from ImageNet database~\cite{deng2009imagenet}. The chosen pictures are cropped into small patches of size 35 $\times$ 35 and Gaussian noise is imposed to these patches. As for the Bergman distance $\Delta_{h}$, we choose Mahalanobis distance $\|\cdot\|_{A}$ as $h$ in our applications~\cite{bauschke1997legendre}.
All experiments are performed on a PC with Intel Core i7 CPU @ 3.4 GHz, 32 RAM and NVIDIA GeForce GTX 1050 Ti GPU.

\subsection{$\ell_p$-Sparse Coding for Image Completion}
We first consider to solve a $\ell_p$-sparse coding model to address the problem of image completion (also known as image inpainting). The purpose of this task is to restore a visually plausible image in which data are missing due to damage or occlusions. This problem can be formulated as:
\begin{equation}
\min\limits_{\bm{\alpha}}\frac{1}{2\rho}\|\mathbf{M}\odot\mathbf{B}\bm{\alpha}-\mathbf{y}\|^2 + \|\bm{\alpha}\|_p^{p},\label{eq:sc}
\end{equation}
where $\mathbf{y} $ is the observed image, $\mathbf{M}$ denotes a mask, $\mathbf{B}$ is the dictionary, $\bm{\alpha}$ is its corresponding sparse coefficients and $\rho>0$ is a parameter. Following \cite{beck2009a}, we consider $\mathbf{B}$ as a inverse wavelet basis (i.e., multiplying by $\mathbf{B}$ corresponds to performing inverse wavelet transform) and thus $\mathbf{B}\bm{\alpha}$ is just the latent image (denoted as $\mathbf{x}$). To enforce the sparsity of $\bm{\alpha}$, we set $A=\mu I$ ($I$ is unit matrix) in Bergman distance and  $p=0.8$ in the above coding model. 

It is easy to check that Eq.~\eqref{eq:sc} is just a specific case of Eq.~\eqref{eq:model} with single variable. In this following, we first verify the theoretical results proposed in this work, and then test the performance of LBS on challenging benchmark datasets.

\textbf{Iteration Behaviors Analysis:}
We first choose example images from CBSD68 dataset~\cite{zhang2017learning} to demonstrate the iterative behaviors of LBS together with some other widely used splitting schemes (e,g, FBS, FISTA, and ADMM).
For fair comparisons, the stopping criterion of all the compared methods are set in the same manner. That is, we denote $\mathbf{x} = \mathbf{B}\bm{\alpha}$ and consider $\|\mathbf{x}^{t+1} -\mathbf{x}^{t}\| / \|\mathbf{x}^{t}\|\leq 10^{-4}$ as the stopping criterion in all these methods.

Fig.~\ref{fig:convergence} showed the convergence curves from different aspects, including iteration error (``$\mathtt{Iter.~Error}$'', defined as $\log(\|\mathbf{x}^{t+1} -\mathbf{x}^{t}\| / \|\mathbf{x}^{t}\| )$) and reconstruction error (``$\mathtt{Rec. ~Error}$'', defined as $\log(\|\mathbf{x}^{t} - \mathbf{x}_{gt}\| / \|\mathbf{x}_{gt}\|)$). Our LBS have superiority against traditional FBS, FISTA, and ADMM on both convergence rates and final reconstruction. LBS only almost a dozen steps can achieve the convergence precision while FBS and FISTA need few hundreds steps and ADMM needs four dozens of steps. 
Since introducing the network as $\mathcal{T}_{d}$, our strategies have lesser reconstruction error than others obviously. The PSNR and SSIM of the final results also verify that our LBS has better performance. Concretely, our PSNR is approximately higher 3dB than the compared methods.

\begin{figure}[t]
	\centering	
	\begin{tabular}{c@{\extracolsep{0.2em}}c}
		\includegraphics[width=0.225\textwidth]{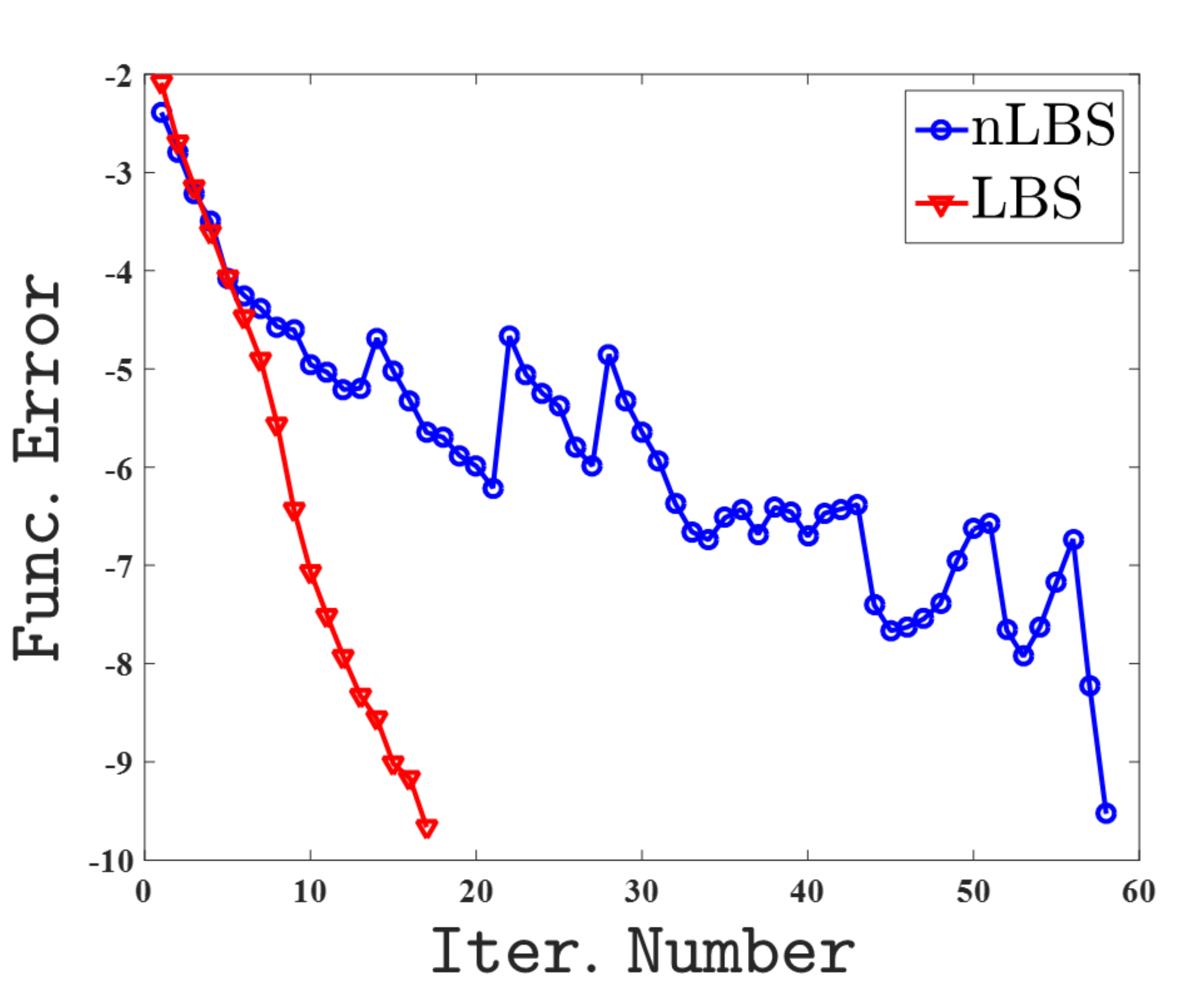}	
		&\includegraphics[width=0.225\textwidth]{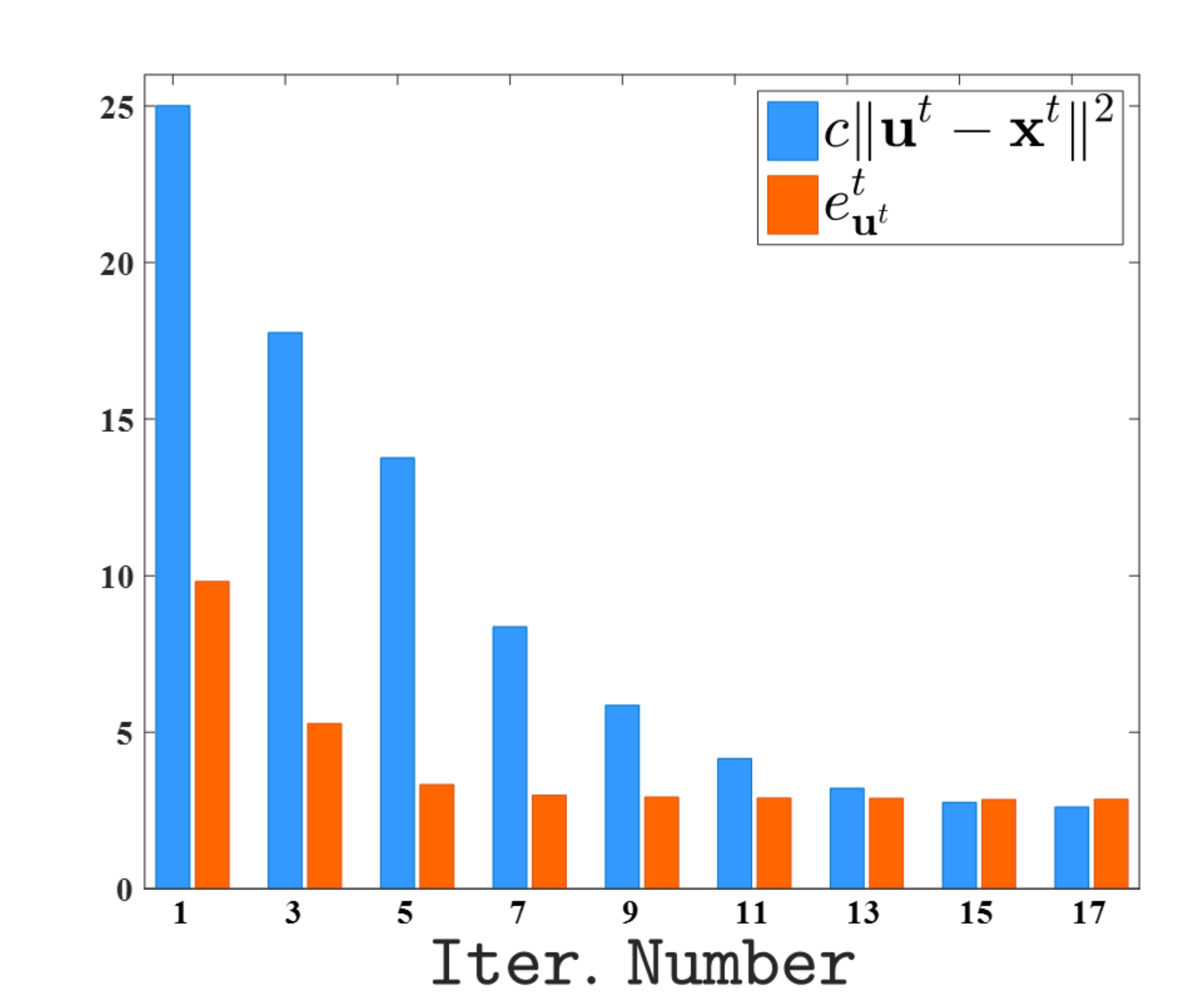}\\
	\end{tabular}
	\caption{The left subfigure plots the ``$\mathtt{Func.~Error}$'' curves of naive LBS (nLBS) and LBS. The right subfigure illustrates when the variables in our algorithm satisfy the ROC criteria during iterations.}
	\label{fig:c_curve}
\end{figure}

We also compared the curves of objective function value errors (``$\mathtt{Func.~Error}$'', based on $\Phi(\mathbf{x}^{t})$) for different settings of LBS, including naive LBS (nLBS, do not check the ROC and monotone conditions) and the complete LBS in Alg.~\ref{alg:LBS}. 
From the left subfigure of Fig.~\ref{fig:c_curve}, it is easy to observe that the proposed criteria can lead to very fast convergence, while there are severe oscillations on the curves of nLBS.
Furthermore, we plotted the bars of ROC (i.e., the error $e_{\mathbf{u}^{t}}^{t}$ and the threshold $c\| \mathbf{u}^{t} - \mathbf{x}^{t}\|^{2}$) on the right part of Fig.~\ref{fig:c_curve}. It can be seen that the ROC condition is always satisfied except at the last two iterations. Thus deep networks are performed at most of our iterations. Only at the last stages, LBS tended to perform model-inspired iterations (i.e., Step~\ref{step:proximal} in Alg.~\ref{alg:LBS}) to obtain accurate solution for the given optimization model.

\begin{table}[!htb]
	\centering
	\caption{Averaged image completion performance with different levels of missing pixels on CBSD68 dataset. The percents of missing pixels are reported in the first column. TIME in the bottom row denotes the averaged run time (in seconds) on all the test images.}\label{table:inpainting} 
	\begin{tabular}{|c|c|c@{\extracolsep{0.2em}}c@{\extracolsep{0.2em}}c@{\extracolsep{0.2em}}c@{\extracolsep{0.2em}}c|}
		\hline
		\%&Metric&FoE&VNL&ISDSB& JSM  &Ours\\
		\hline
		\multirow{2}{*}{20}
		&PSNR&38.23&28.87&35.20&37.55  &\textbf{38.77}\\ 
		&SSIM&0.95&0.95&0.96&0.98  &\textbf{0.98}\\ \cline{1-7}
		\multirow{2}{*}{40}
		&PSNR &34.01&27.55&31.32&33.54  &\textbf{34.54}\\ 
		&SSIM &0.90&0.91&0.91& 0.94 &\textbf{0.95}\\ \cline{1-7}
		\multirow{2}{*}{60}
		&PSNR &30.81&26.13&28.23&29.96   &\textbf{31.27}\\ 
		&SSIM &0.81&0.85&0.83& 0.81  &\textbf{0.90}\\ \cline{1-7}
		\multirow{2}{*}{80}
		&PSNR &27.64&24.23&24.92& 27.32   &\textbf{27.71}\\ 
		&SSIM &0.65&0.75&0.70& 0.79  &\textbf{0.80}\\
		\hline
		-&TIME&34.85 &1515.49 &28.00 &207.57 &\textbf{1.40}\\ \hline
	\end{tabular}
	\label{tab:datasetresults}
\end{table}

\textbf{Comparisons on Benchmarks:}
To further express the superiority of LBS, we generated random masks of different levels (including 20\%, 40\%, 60\% and 80\% missing pixels) on CBSD68 dataset~\cite{zhang2017learning} for comparison, which contains 68 images with the size of 481$\times$321. 
Then we compared LBS with four state-of-the-art methods, namely, FoE~\cite{Roth2009Fields}, VNL~\cite{Arias2011A}, ISDSB~\cite{He2014Iterative}, and JSM~\cite{zhang2014image}. Tab.~\ref{table:inpainting} reports the averaged quantitative results, including PSNR, SSIM, and time (in second). It can be seen that regardless the proportion of masks, LBS can achieve better performance against the state-of-the-art approaches. This is mainly due to our superior strategy which using learnable network operator. 

\begin{figure*}[t]
	\centering	\begin{tabular}{c@{\extracolsep{0.2em}}c@{\extracolsep{0.2em}}c@{\extracolsep{0.2em}}c@{\extracolsep{0.2em}}c@{\extracolsep{0.2em}}c}
		\includegraphics[width=0.16\textwidth]{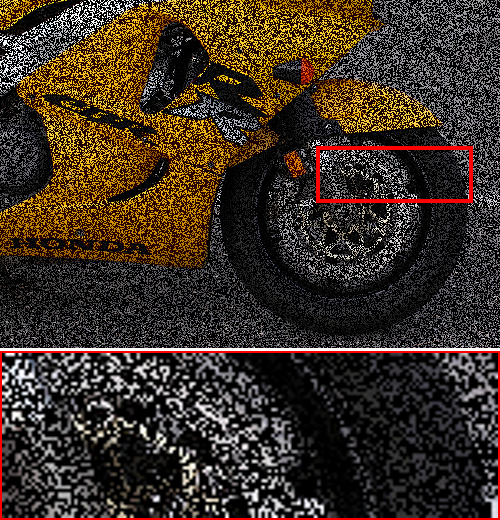}
		&\includegraphics[width=0.16\textwidth]{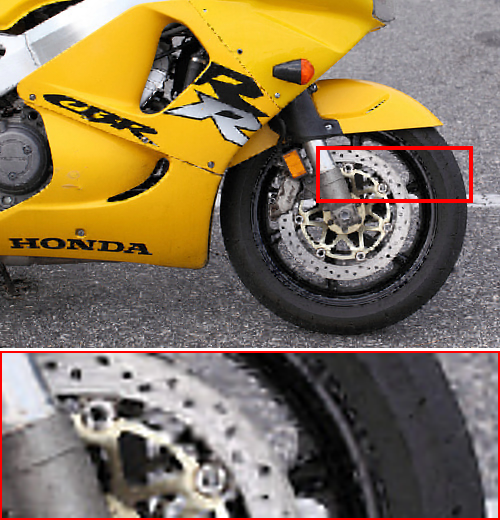}
		&\includegraphics[width=0.16\textwidth]{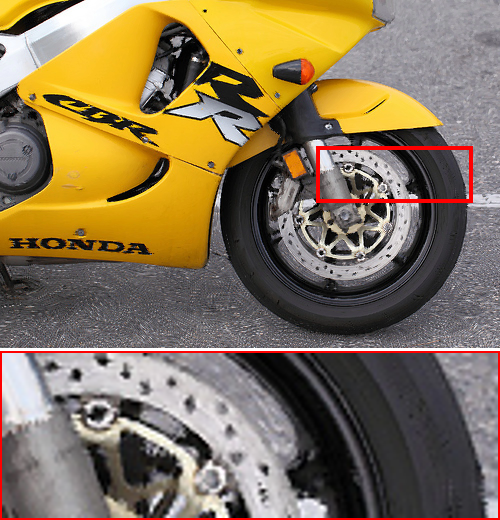}
		&\includegraphics[width=0.16\textwidth]{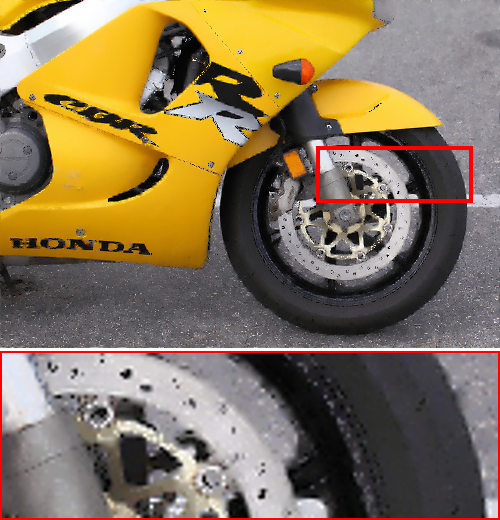}
		&\includegraphics[width=0.16\textwidth]{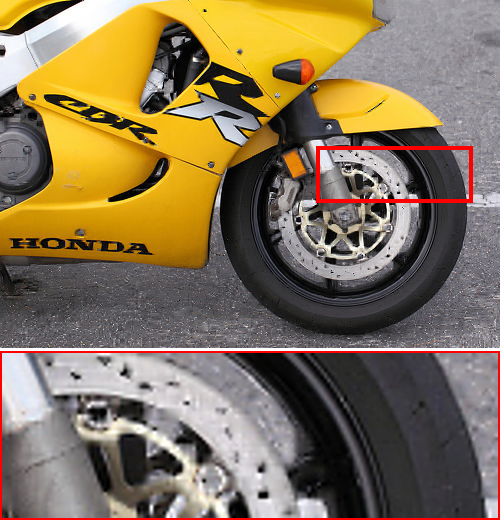}
		&\includegraphics[width=0.16\textwidth]{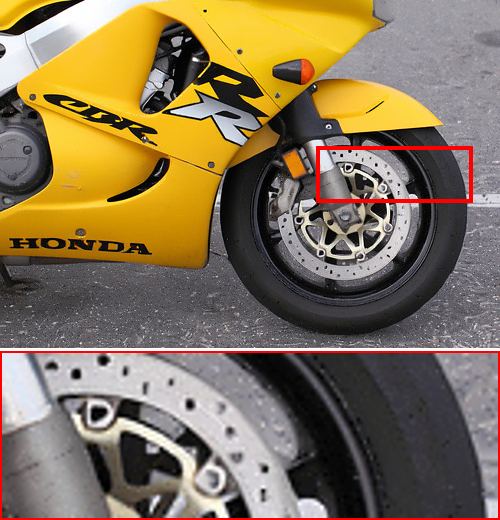}\\
		Input &  FoE  &VNL   & ISDSB  &JSM  & LBS  \\	
		-& (24.45 / 0.86)&(24.92 / 0.86)&(23.27 / 0.83)&(25.40 / 0.87) &(\textbf{26.11} / \textbf{0.88})\\
	\end{tabular}
	\caption{The inpainting results (with 60\% missing pixels) of LBS with comparisons to state-of-the-art methods. The quantitative scores (i.e., PSNR / SSIM) are reported below each image. }
	\label{fig:inpaint-com}
\end{figure*}

We then compared the visual performance of LBS with all these methods. Fig.~\ref{fig:inpaint-com} presented the comparisons on an image from ImageNet database~\cite{deng2009imagenet} with 60\% missing pixels. It can be seen that LBS outperformed all the compared methods on both visualization and metrics (PSNR and SSIM). The edge of motorcycle wheels can be restored more smooth and clear by LBS, while other approaches exist some noises and masks to affect the visual effects.

\subsection{Nonconvex TV for Image Deblurring}
We further evaluate LBS on image deblurring, which is a challenging problem in computer vision area.
Here we consider the following widely used total variation (TV) based formulation:
\begin{equation}
\min\limits_{\mathbf{u}}\frac{1}{2\rho}\|\mathbf{k}\otimes\mathbf{u}-\mathbf{y}\|^2 + \mathtt{TV}_p(\mathbf{u})  + \chi_{\Omega_{\mathbf{u}}}(\mathbf{u}) ,
\label{eq:deblur-model}
\end{equation}
where $\mathbf{k}, \mathbf{x}, \mathbf{y}$ denote the blur kernel, latent image, and blurry observation, respectively. $\mathtt{TV}_p(\mathbf{u})=\|\mathbf{D}_h\mathbf{u}\|_p^p+\|\mathbf{D}_v\mathbf{u}\|_p^p$ is the nonconvex TV regularization with gradient matrices $\mathbf{D}_h$ and $\mathbf{D}_v$ (here we also set $p=0.8$ for the $\ell_p$ norm). $\chi_{\Omega_{\mathbf{u}}}(\mathbf{u}) $ is the indicator function of the set $\Omega_{\mathbf{u}} := \{\mathbf{u} \in \mathcal{R}^{n}: 0 \leq \mathbf{u}_{i}\leq 1 \}.$
Following the half-quadratic splitting technique, Eq.~\eqref{eq:deblur-model} (with auxiliary variables $\mathbf{v}_h$ and $\mathbf{v}_v$) can be reformulated as 
\begin{equation}
\begin{array}{l}
\min\limits_{\mathbf{u},\mathbf{v}_h,\mathbf{v}_v}\frac{1}{2\rho}\|\mathbf{k}\otimes\mathbf{u}-\mathbf{y}\|^2 
+  \|\mathbf{v}_h\|_p^p+\|\mathbf{v}_v\|_p^p\\ + \chi_{\Omega_{\mathbf{u}}}(\mathbf{u})  
+ \frac{1}{2\eta}\left(\|\mathbf{D}_h\mathbf{u}-\mathbf{v}_h\|^2+\|\mathbf{D}_v\mathbf{u}-\mathbf{v}_v\|^2\right).
\end{array}\label{eq:ntv}
\end{equation}
Obviously, Eq.~\eqref{eq:ntv} is a special case of Eq.~\eqref{eq:model} with three blocks. Thus can be efficiently addressed by LBS. We adopt $A = \text{diag}\left(\mu_{\mathbf{u}} I_{\mathbf{u}}, \mu_{\mathbf{v}_h} I_{\mathbf{v}_h}, \mu_{\mathbf{v}_v} I_{\mathbf{v}_v}\right)$ in Bergman distance $\Delta_h$ which satisfies $I_{\{\cdot\}}$s are unit matrices, $\mu_{\mathbf{u}} = 0.01$, and $\mu_{\mathbf{v}_h}=\mu_{\mathbf{v}_v} = 0.001$.
\begin{figure}[t]
	\centering	\begin{tabular}{c}
		\includegraphics[width=0.46\textwidth]{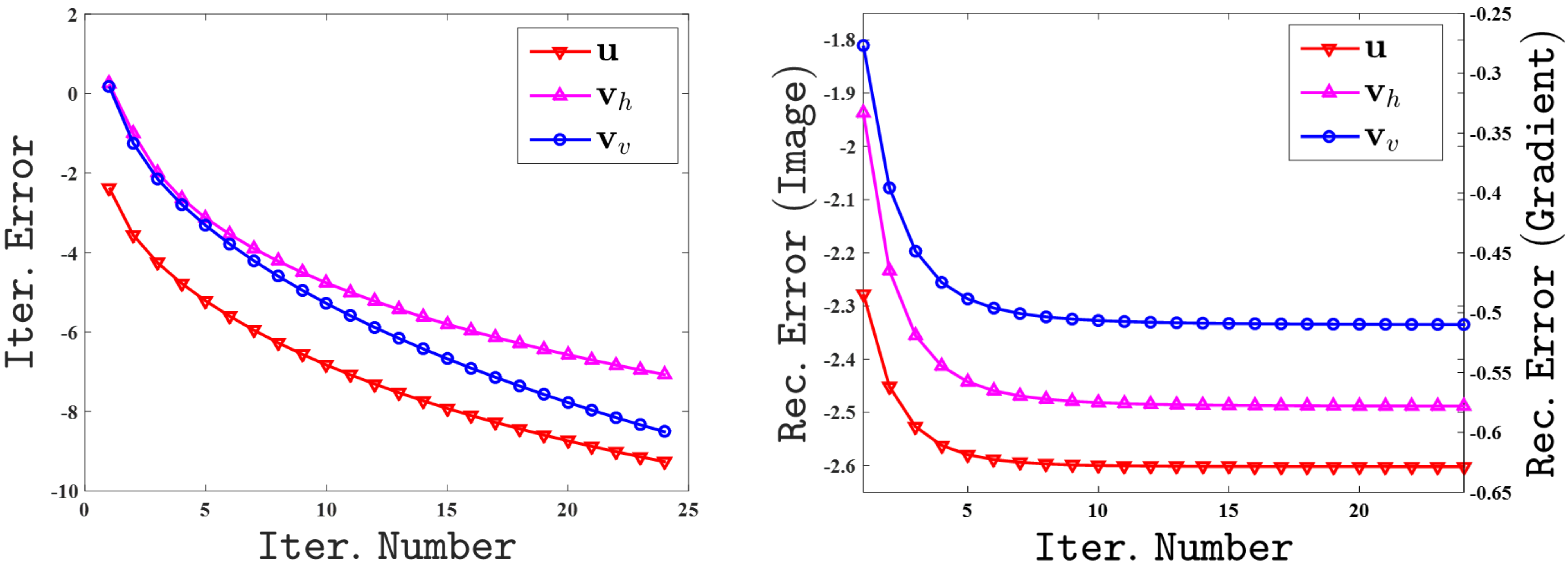}	\\
	\end{tabular}
	\caption{The iteration curves of the multiple variables in LBS. In the right subfigure, the left and right vertical ordinates are the $\mathtt{Rec.~Error}$ of image (i.e., $\mathbf{u}$) and gradients (i.e., $\mathbf{v}_h$ and $\mathbf{v}_v$) respectively.
	}
	\label{fig:multi-convergence}
\end{figure}
\begin{figure*}[t]
	\centering	\begin{tabular}{c@{\extracolsep{0.2em}}c@{\extracolsep{0.2em}}c@{\extracolsep{0.2em}}c@{\extracolsep{0.2em}}c@{\extracolsep{0.2em}}c}	
		\includegraphics[width=0.16\textwidth]{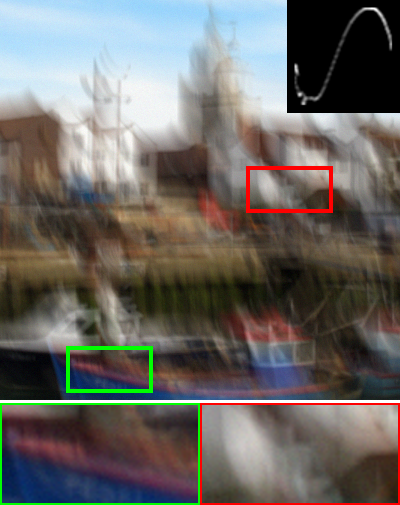}
		&\includegraphics[width=0.16\textwidth]{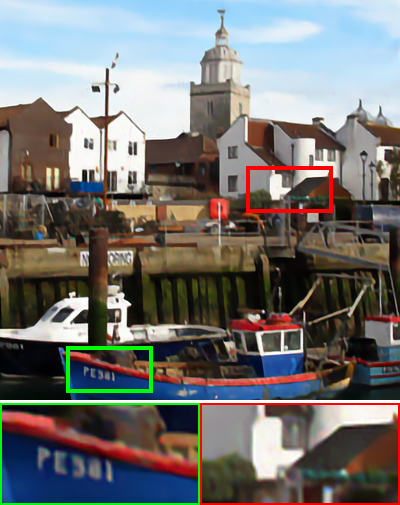}
		&\includegraphics[width=0.16\textwidth]{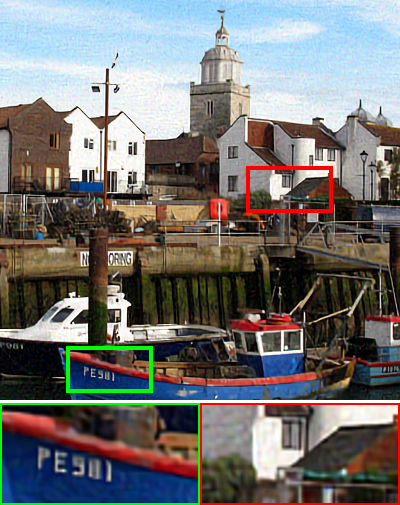}
		&\includegraphics[width=0.16\textwidth]{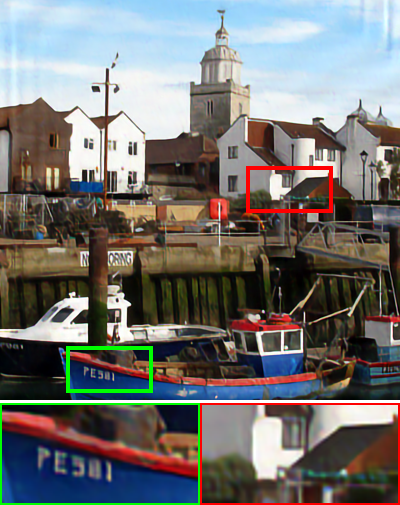}
		&\includegraphics[width=0.16\textwidth]{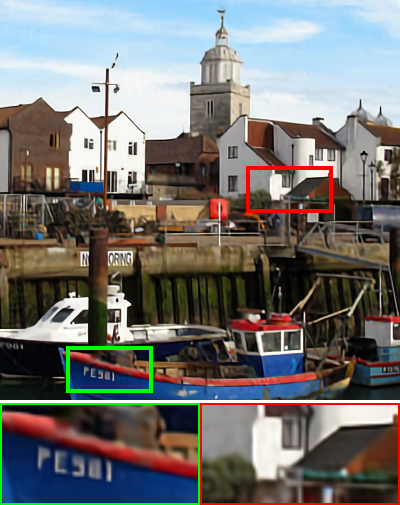}
		&\includegraphics[width=0.16\textwidth]{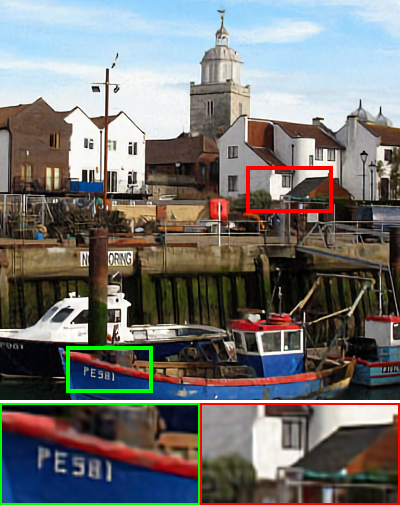}\\
		Input & RTF   & IRCNN  & FDN   & PADNet& LBS  \\
		- &(22.78 / 0.75) & (25.00 / 0.77) &(23.79 / 0.78) & (24.39 / 0.81) &(\textbf{25.34} / \textbf{0.83})\\
	\end{tabular}
	\caption{The deblurring results (by a large scale blur kernel with the size 75$\times$75) of LBS with comparisons to state-of-the-art methods. The quantitative scores (PSNR / SSIM) are reported below each method. }
	\label{fig:deconvolution-com}
\end{figure*}
\begin{table*}[!htb]
	\centering
	\caption{Averaged quantitative results of image deblurring on the Sun $et \ al.$ benchmark image set.}
	\begin{tabular}{|c|c|c|c|c|c|c|c|c|c|c|c|}
		\hline
		Metric&TV&HL&CSF&IDDBM3D&EPLL&RTF&MLP&IRCNN&FDN&PADNet&Ours\\ \hline
		PSNR&30.67&31.03&31.55&30.79&32.44&32.45&31.47&32.61&32.65&32.69&\textbf{32.90} \\ 
		SSIM&0.85&0.85&0.87&0.87&0.88&0.89&0.86&0.89&0.89&0.89&\textbf{0.90} \\ \cline{1-12}
	\end{tabular}
	\label{table:non-blind-dataSetResult}
\end{table*}

Fig.~\ref{fig:multi-convergence} demonstrated the convergence behaviors of LBS on $\{\mathbf{u}, \mathbf{v}_{h},\mathbf{v}_{v}\}$. It can be seen from the left subfigure that ``$\mathtt{Iter.~Error}$'' of all blocks quickly decreased to $ \log(10^{-3})$, notice that ``$\mathtt{Iter.~Error}$'' of $\mathbf{u}$ is even less than $\log(10^{-4})$. On the right subfigure, the ``$\mathtt{Rec. ~Error}$'' of $\mathbf{v}_{h}$ and $\mathbf{v}_{v}$ also have dramatic decline trend, which are shown along with the right vertical ordinate. Due to the different range of values, we have to plot the curves of $\mathbf{u}$ w.r.t. the left vertical ordinate. We can see that it still obtained the least ``$\mathtt{Rec. ~Error}$''. 

We then reported results on the challenging image deblurring benchmark dataset collected by Sun $et \ al.$~\cite{sun2013edge} (which includes 640 blurry images with 1\% Gaussian noises) for quantitative evaluation. We compared LBS with plenty of competitive approaches, including TV~\cite{wang2008a}, HL~\cite{krishnan2009fast}, CSF~\cite{schmidt2014shrinkage}, IDDBM3D~\cite{danielyan2012bm3d}, EPLL~\cite{zoran2011learning}, RTF~\cite{schmidt2016cascades}, MLP~\cite{schuler2013machine}, IRCNN~\cite{zhang2017learning}, FDN~\cite{Kruse2017Learning}, and PADNet~\cite{liu2018proximal}). 

It is known that learning-based methods (e.g., CSF, RTF, MLP, IRCNN, FDN, and PADNet) can achieve better performance than other conventional approaches in terms of quantitative metrics (e.g., PSNR and SSIM). 
However, due to the weak theoretical guarantee, they are worse than LBS (see Tab.~\ref{table:non-blind-dataSetResult}).
Fig.~\ref{fig:deconvolution-com} expressed the qualitative results of LBS against other methods (top 4 in Tab.~\ref{table:non-blind-dataSetResult}) on an example blurry image, which is generated with a large scale blur kernel (75$\times$75 pixels) on an image from ImageNet~\cite{deng2009imagenet}. It can be seen that LBS can restore the text and windows more distinctly than others. Although IRCNN has relatively higher PSNR than others (but lower than LBS), its visual quality and SSIM are not satisfied.

\section{Conclusions}
This paper proposed Learnable Bregman Splitting (LBS), a novel deep architectures based operator splitting algorithm for task-specific nonconvex optimization. It is demonstrated that both the model-based operators and the data-dependent networks can be used in our iteration. We also provided solid theoretical analysis to guarantee the convergence of LBS. The experimental results verified that LBS can obtain better performance against most other state-of-the-art approaches. 
\section{Acknowledgments}
This work is partially supported by the National Natural Science Foundation of China (Nos. 61672125, 61733002, 61572096, 61432003 and 61632019), and the Fundamental Research Funds for the Central Universities.

\bibliographystyle{named}
\bibliography{lbs}

\end{document}